\documentclass[11pt]{article}

\usepackage[utf8]{inputenc}
\usepackage[T1]{fontenc}
\usepackage{lmodern}
\usepackage{graphicx}
\usepackage{amsmath, amssymb}
\usepackage{booktabs}
\usepackage{hyperref}
\usepackage{microtype}
\usepackage{cite}
\usepackage{geometry}
\title{Insights on back marking for the automated identification of animals}
\author{
  D. Brunner$^{1, 2, \dagger}$,
  M. Bordes$^{3}$,
  E. Mayrhuber$^{1, 2}$,
  S. M. Winkler$^{1}$,
  V. Dorfer$^{1}$
  and 
  M. Oczak$^{3, 4}$
}

\begin{document}
\date{}
\maketitle
\vspace{-24pt}

\begin{center}
  $^{1}$\textit{Bioinformatics Research Group, PLFDoc, University of Applied Sciences Upper Austria, Hagenberg, Austria} \\

  $^{2}$\textit{Computer Vision Lab, TU Wien, Vienna, Austria} \\

  $^{3}$\textit{Centre for Animal Nutrition and Welfare, Farm Animals and Food System Transformation, University of Veterinary Medicine Vienna, Vienna, Austria} \\

  $^{4}$\textit{Precision Livestock Farming Hub, University of Veterinary Medicine Vienna, Vienna, Austria} \\

  $^{\dagger}$david.brunner@fh-hagenberg.at \\
\end{center}

\vspace{12pt}

\begin{abstract}
To date, there is little research on how to design back marks to best support individual-level monitoring of uniform looking species like pigs. With the recent surge of machine learning-based monitoring solutions, there is a particular need for guidelines on the design of marks that can be effectively recognised by such algorithms. This study provides valuable insights on effective back mark design, based on the analysis of a machine learning model, trained to distinguish pigs via their back marks. Specifically, a neural network of type ResNet-50 was trained to classify ten pigs with unique back marks. The analysis of the model’s predictions highlights the significance of certain design choices. Most importantly, the set of back marks must be designed such that each mark remains unambiguous under conditions of motion blur, diverse view angles and occlusions, caused by animal behaviour. The generated insights can support individual-level monitoring in future studies and real-world applications by optimizing back mark design.
\end{abstract}


\section{Introduction}
Observing animals is a well-established way of gathering information about their social behaviour \cite{clouard2024evidence}. While, traditionally, this is done by a human expert, in recent years, motivated by advancements in machine learning (ML), much research has been devoted to the development of automatic monitoring solutions. A plethora of studies show ML to be capable of accurately detecting, tracking, and ultimately recognising the behaviour of animals \cite{liu2020computer}. One obstacle towards practical application is the difficulty of identifying individual animals. This is in large part because differentiating individual animals is a very hard problem in general, and especially so for uniform looking species like pigs. One common solution is to use additional physical markers like back marks. To date, there are only few studies that provide insights on their design \cite{ohayon2013automated}. In this work, an existing study is considered in which back marks were designed for pigs based on the limited information available. To analyse the back mark’s suitability for automated identification, a machine learning model is trained to recognise pigs via these back marks. The analysis of the model’s performance unveils interesting error cases and allows deriving design recommendations for back marks in future studies.

\section{Material and Methods}
\begin{figure}[h]
\centering
\includegraphics[width=1.0\linewidth]{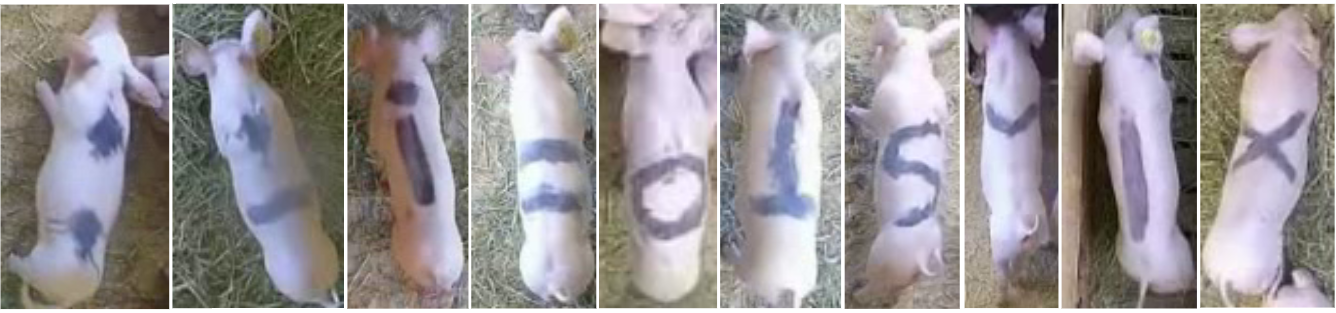}
\caption{The back marks used in this study. From left to right: \textit{dot dot}, \textit{dot line horizontal}, \textit{i}, \textit{line line horizontal}, \textit{o}, \textit{reverse t}, \textit{s}, \textit{v}, \textit{vertical line}, \textit{x}.}
\label{fig:backmarks}
\end{figure}

The data was collected during an observational study on social behaviour in pigs, focused on helping behaviour (“Let me out”, doi:10.55776/I6488). The study setup comprised two identical pens, observed by identical cameras mounted in an elevated side-view angle. The pigs received regular back marks (1-2 times/week). Symbols were used instead of numbers for practicality (easy to apply to moving pigs) and reproducibility (similar across reapplications). The back marks were inspired by patterns used in mice \cite{ohayon2013automated} and are shown in Fig. \ref{fig:backmarks}. For the dataset creation, a human labeller placed bounding boxes with class IDs on all frames of 11 video clips, each 10-30 seconds in duration. Nine of the clips were selected for the training set, and two of the clips for the validation and test set, respectively. The test clip was deliberately chosen to constitute an especially challenging setting (fast movement, occlusions). To combat the high inter-frame redundancy inherent to videos, the validation and test videos were sampled at a low framerate (5 fps). The training videos were sampled at the original framerate (24 fps), because extensive data augmentation (flipping, rotation, colour jitter, grayscale, blurring) allowed making use of near-duplicate frames in the training set, significantly extending it in size. The bounding box areas were then algorithmically cropped from all frames and manually filtered. The final training set comprised 26,260 crops, the validation and test set 500 crops each. A PyTorch implementation of a neural network of type ResNet-50 \cite{he2016deep} was selected as the image classification model and trained for 150 epochs.

\section{Results and Discussion}
\begin{figure}[h]
\centering
\includegraphics[width=0.9\linewidth]{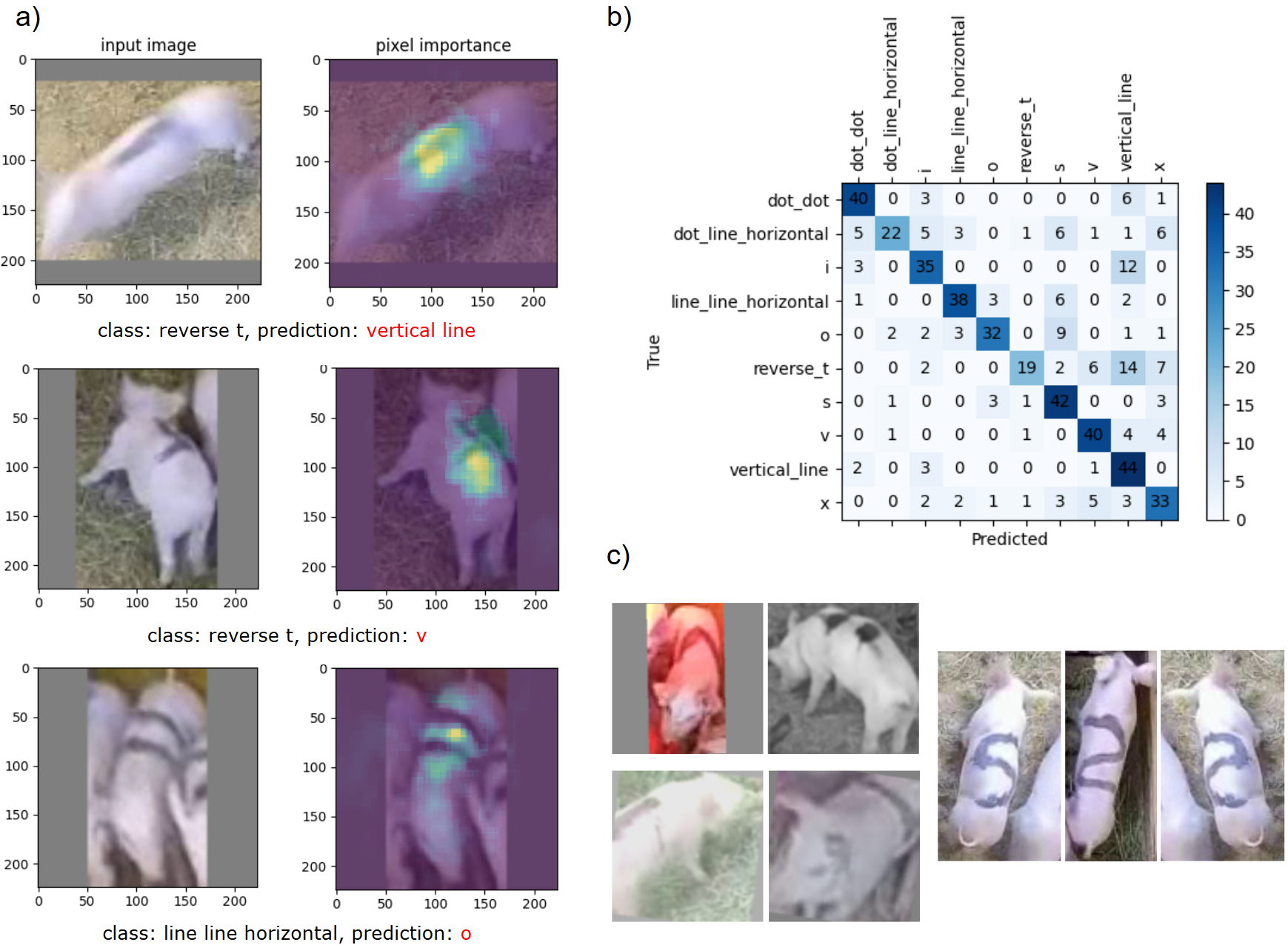}
\caption{a) Behaviour-based error cases, from top to bottom: motion blur, angle, occlusion. b) Test set confusion matrix. c) Augmentations also can introduce ambiguity.}
\label{fig:errors-conf-aug}
\end{figure}

Table \ref{tab:class_acc} compares the classification accuracy on validation and test set. Fig. \ref{fig:errors-conf-aug}b gives details on the test set performance and reveals conspicuous differences across classes. Especially \textit{reverse t} poses a challenge for the model, which might be explained by the fact that it is prone to look similar to other back marks in specific situations, which are often tied to specific types of animal behaviour. In this study, three kinds of behaviour were identified as especially relevant: fast movement, diverse poses and proximity to other animals, which can lead to motion blur, diverse view angles and occlusions, respectively. The top line in Fig. \ref{fig:errors-conf-aug}a illustrates how motion blur can cause \textit{reverse t} to look like \textit{vertical line}. The pixel importance \cite{zeiler2014visualizing} shows that the blurred vertical bar of the T does not factor into the model’s decision. The middle line in Fig. \ref{fig:errors-conf-aug}a illustrates how a certain view angle can cause \textit{reverse t} to look more like \textit{v} or \textit{x}. The bottom line of Fig. \ref{fig:errors-conf-aug}a shows how even minor occlusions can make it impossible for the model to distinguish between \textit{line line horizontal} and \textit{o}. Back marks must be chosen, such that these situations are avoided. Additional recommendations pertain to data augmentation. Fig. \ref{fig:errors-conf-aug}c illustrates how colour jitter and grayscale (top left pair) might interfere with coloured back marks, how cropping (bottom left pair) makes it impossible to differentiate certain pairs of marks (\textit{reverse t} vs. \textit{vertical line}, \textit{line line horizontal} vs. \textit{o}), and flipping (right triplet) introduces ambiguity in the case of back marks that are mirror images of one another (e.g., \textit{s} and \textit{2}).

\begin{table}[ht]
    \centering
    \caption{The classification accuracy on the validation and test sets.}
    \label{tab:class_acc}
    \begin{tabular}{lr}
        \toprule
        \textbf{class} & \textbf{acc. val. / test (\%)} \\
        \midrule dot dot & 88 / 80 \\
        dot line horizontal & 100 / 44 \\
        i & 98 / 70 \\
        line line horizontal & 100 / 76 \\
        o & 82 / 64 \\
        reverse t & 66 / 38 \\
        s & 76 / 84 \\
        v & 100 / 80 \\
        vertical line & 100 / 88 \\
        x & 100 / 66 \\
        total & 91 / 69 \\
        \bottomrule
    \end{tabular}
\end{table}

\clearpage
The main contribution of this work is the insight that back mark design has to take into account animal behaviour, which has not been discussed in previous work \cite{kashiha2013automatic}. Animals move in certain ways (e.g., frantic sprints), take on certain poses (e.g., lying on the side) and show characteristic social behaviour (e.g., huddling together). As shown in Fig. \ref{fig:errors-conf-aug}a, these behaviours can introduce ambiguity into the appearance of the back marks. Since the presented analysis was conducted only after the observational study had already moved into the experimental phase, the generated insights could not be used or evaluated within this study. Instead, the insights serve as recommendations for future studies, which are tasked with designing marks that are robust to the described scenarios.

\section{Conclusions}
This study aimed at investigating back mark design for identification tasks in automated animal monitoring. It could be shown that there are important design choices that impact a ML model’s ability to successfully differentiate a set of back marks. Specifically, the back mark design should control for predictable animal behaviour such as fast movement, diverse poses and proximity to other animals.

\section*{Acknowledgements}
This research was funded in whole or in part by the Austrian Science Fund (FWF) [\url{https://doi.org/10.55776/DFH34}]. For open access purposes, the author has applied a CC BY public copyright license to any author-accepted manuscript version arising from this submission. The data used in this study originates from the "Let me out" project, funded by the Austrian Science Fund (FWF) [\url{https://doi.org/10.55776/I6488}].

\bibliographystyle{plain}
\bibliography{references}

\end{document}